\documentclass[sigconf]{acmart}

\setcopyright{none}
\acmConference{}{}{}
\acmBooktitle{}
\renewcommand\footnotetextcopyrightpermission[1]{}

\usepackage{booktabs}
\usepackage{array}

\newcolumntype{L}[1]{>{\raggedright\arraybackslash}p{#1}}
\newcommand{\rot}[1]{\rotatebox{90}{#1}}
\begin{document}

\title{SupplyChain-JEPA: Pretraining a Predictive World Model for Supply Chain Operations}

\author{Haoting Zhang}
\email{haoting_zhang@berkeley.edu}
\affiliation{\institution{University of California, Berkeley}\city{Berkeley}\state{CA}\country{USA}}

\author{Yunduan Lin}
\email{yunduanlin@cuhk.edu.hk}
\affiliation{\institution{The Chinese University of Hong Kong}\city{Hong Kong}\country{China}}

\author{Donglin Zhan}
\email{dz2478@columbia.edu}
\affiliation{\institution{Columbia University}\city{New York}\state{NY}\country{USA}}

\author{Tong Wang}
\authornote{Corresponding author.}
\email{twang533@cityu.edu.hk}
\affiliation{\institution{City University of Hong Kong}\city{Hong Kong}\country{China}}

\author{Chung-Piaw Teo}
\email{bizteocp@nus.edu.sg}
\affiliation{\institution{National University of Singapore}\country{Singapore}}

\begin{abstract}
Supply chain decisions depend on operating conditions that records show in part. Stockouts censor demand, tracking reports arrive late, and order logs record the plan that was executed. Within a specific application, recovering the missing information is costly, and logs do not contain the outcomes of actions that were not taken. Other processes and simulators often record what a specific application hides. Combining these sources calls for one model. We propose SupplyChain-JEPA, a predictive world model pretrained on operational datasets and simulators. A common record format, which uses concepts from operations management, stores each value with its observation status and the time it became available. Following the joint-embedding predictive architecture, the State component learns to predict withheld current and future information under masks that follow real information gaps. The Dynamics component learns from paired simulator branches how the state changes under a candidate order. We apply our trained SupplyChain-JEPA model to three decisions: demand recovery under stockouts, early ranking of shipment risk, and replenishment-plan choice. Pretraining excludes the data of these decisions. On each decision, our system performs better than a conventional baseline.
\end{abstract}


\maketitle
\pagestyle{plain}

\begin{figure*}[t]
\centering
\includegraphics[width=\textwidth]{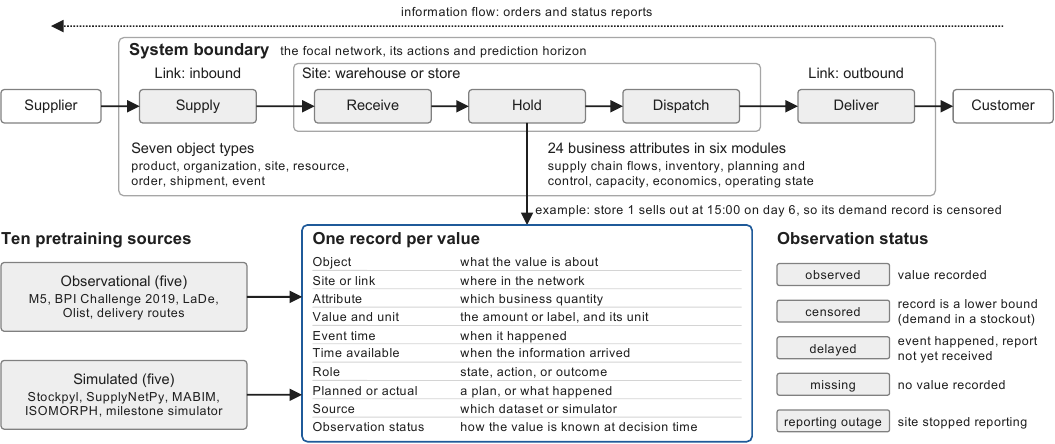}
\caption{The operating system and the common record. Material moves from the supplier to the customer in five steps, over links and at a site. Orders and status reports form the information flow (dotted arrow). Seven object types and 24 business attributes describe the system. Each value becomes one record, and the center panel gives the meaning of each component. In the example, store 1 sells out at 15:00 on day 6, so its sales give a lower bound on demand. Delay labels can be assigned retrospectively (Table~\ref{tab:records}).}
\Description{Material flows through a focal site. The diagram maps objects and operational attributes to records and shows the ten sources.}
\label{fig:representation}
\end{figure*}

\section{Introduction}

Supply chain managers place orders, allocate inventory, and select shipments for review before the information they need is complete. Consider two stores that each sell 100 units of a product in one day. One store sells out in the afternoon, and the other still has stock at closing. Under a lost-sales model with no substitution, the second store's sales reveal realized demand, while the first store's sales give only a lower bound. An earlier stocking decision changes what the same sales figure tells the planner. In freight operations, a missing tracking event can mean a late shipment or a late report. Inventory records can also differ from physical stock. An audit of nearly 370,000 inventory records across 37 stores of one retailer found about 65\% of them inaccurate~\cite{dehoratius2008inaccuracy}. Managers therefore need to infer, from the records they observe, the state on which the next decision depends.

Operations management offers methods for each of these gaps within a specific application. Censored-demand methods learn demand from sales and adapt orders as data accumulate~\cite{huh2011censored}. Bayesian inventory models update a belief about physical stock from sales and replenishment records~\cite{dehoratius2008management}. In logistics, predictive process monitoring uses the observed part of a running process to anticipate its outcome~\cite{metzger2015predictive}.

Application-specific evidence can be expensive. Recording stockout times helps recover profit lost to censoring~\cite{jain2015stockout}, while learning from sales alone can require holding extra stock to observe demand~\cite{besbes2013censoring}. Historical order logs contain outcomes only for the executed plans.

Other processes and simulators often record what a specific application hides. A simulator records true demand next to censored sales, and it can run two plans from the same state under the same random demand. Other datasets record event timing or route deviations that a specific application lacks. When many problems each have little data, pooling data across them can lower their total cost~\cite{gupta2022pooling}. These sources, however, differ in their fields, time scales, and patterns of missing data. The question is whether relationships observed in one process can help interpret partial records in another when the firms and episodes differ.

We study a world model that summarizes the available history into a state and predicts how this state evolves under candidate actions~\cite{hafner2019planet}. In partially observed control, a summary that predicts observations and costs can support action choice~\cite{subramanian2022ais}. Pretrained time-series models transfer forecasts across datasets~\cite{das2024timesfm}. Here, transfer also requires preserving when a report became available and which action generated an outcome. Each application can then fit its own estimator or decision rule.

To answer this question, we propose SupplyChain-JEPA. It adopts the learning principle of the joint-embedding predictive architecture (JEPA)~\cite{assran2023ijepa}, which predicts a representation of withheld information from the visible information. Versions of this principle exist for video with actions~\cite{assran2025vjepa2}, tabular data~\cite{thimonier2025tjepa}, and time series~\cite{verdenius2024latpfn}. In images, withholding is a training device, because the full image is available when the model is used. In supply chains, missing information is the problem itself. We therefore withhold information with masks that follow real information gaps, such as stopped reports, outages, unreported milestones, and future periods. Because the targets are learned representations, sources with different fields share one prediction task. The model also predicts business quantities, such as demand and lead time, so that the representation keeps the meaning and magnitude that decisions use. Our main contributions are summarized as follows.

\begin{enumerate}
\item We propose a common record format for incomplete operational data (Section~\ref{sec:representation}). It describes the operating system with seven object types and 24 business attributes from operations management. It stores each value with its observation status and the time it became available.
\item We employ a two-stage training procedure (Section~\ref{sec:training}). State pretraining predicts withheld current and future information under masks that follow real information gaps. Dynamics training then uses paired simulator branches under common random numbers to learn how two orders change the future state.
\item We apply our trained SupplyChain-JEPA model to three decisions whose data are excluded from pretraining (Section~\ref{sec:experiments}). On the primary evaluations, the system reduces error on masked in-stock sales by 4.28\%, raises shipment-risk AUPRC by 6.74\% on the confirmation cohort, and lowers mean replenishment regret by 36.6\%, each relative to its task-specific baseline.
\end{enumerate}
For an e-commerce planning assistant, the resulting model could provide numerical estimates and plan comparisons from the records available at decision time.

\section{SupplyChain-JEPA}
\label{sec:method}

The model first learns a state from incomplete records, then learns how that state evolves under an order.

\subsection{From the operating system to records}
\label{sec:representation}

\begin{figure*}[t]
\centering
\includegraphics[width=\textwidth]{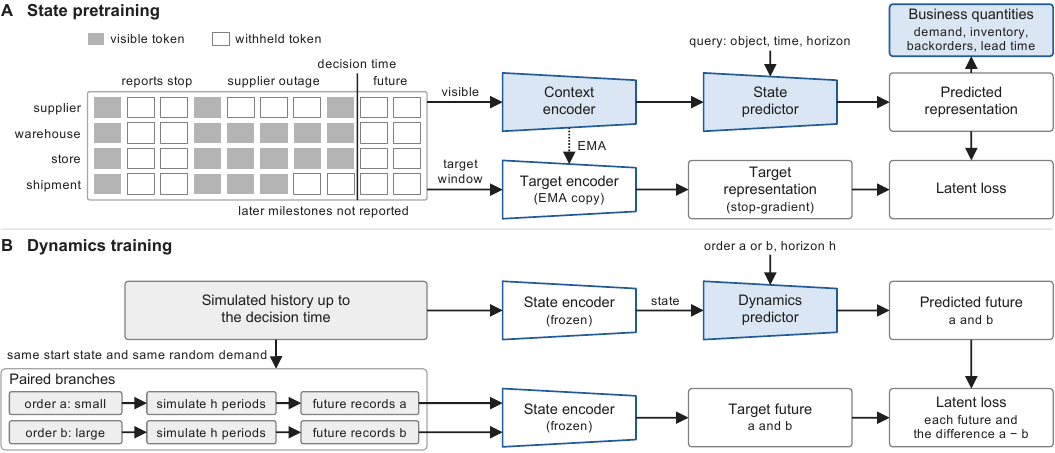}
\caption{Two-stage training. (A) State pretraining. Masks that follow real information gaps withhold parts of an episode. The grid shows four mask types. The target encoder, an exponential moving average (EMA) of the context encoder, reads the unmasked target window and receives no gradient. The loss uses its outputs at the withheld tokens. A head trained in this stage predicts business quantities from the State predictor's output. (B) Dynamics training. The State encoder stays frozen. Two simulator branches start from the same state, use the same random demand, and differ in the order. Each stage trains the blue blocks.}
\Description{The State encoder learns from masked episodes. A frozen copy encodes the paired simulator futures used in Dynamics training.}
\label{fig:training}
\end{figure*}

We describe the data with concepts from operations management~\cite{slack2022operations,chopra2019supply}. The description starts with a system boundary: the focal network, the activities it includes, the actions available to managers, and the prediction horizon. Within this boundary, material moves from suppliers to customers in five steps: supply, receive, hold, dispatch, and deliver (Figure~\ref{fig:representation}). Supply and delivery use inbound and outbound links. The other three steps occur at a site, such as a warehouse or a store. Orders and status reports form the information flow. Seven object types describe the system: product or material, organization, site, resource, order, shipment, and event. For example, an order requests a product, a shipment carries it along a link, and a site holds inventory of it. We use two examples in this section: a store that sells out in the afternoon, and a shipment whose milestone report is late.

Twenty-four business attributes describe these objects. They form six modules that follow textbook topics: supply chain flows, inventory, planning and control, capacity, economics, and operating state. For the store, the flows module includes demand and sales, and the inventory module includes on-hand stock and a stockout indicator. For the shipment, the flows module includes delivery deviation, and the planning and control module includes the execution status of each milestone. Operating state has one attribute: the operating regime or availability of a site or resource, such as normal or disrupted. Totals, counts, fill rates, and net stock are derived quantities with explicit formulas. Attributes that look alike keep different meanings. Demand differs from sales during a stockout, and on-hand stock differs from net stock when there are backorders.

Records describe this system one value at a time. Each record names the object, the site or link, the attribute, and the value with its unit. It gives the time of the event and the time the information became available. It also states the role of the value, such as state, action, or outcome, and whether the value is planned or actual. Finally, it names the source and an observation status: observed, censored, delayed, missing, or reporting outage. The format keeps an unrecorded value distinct from a recorded zero.

\begin{table}[t]
\caption{Records for the two examples, one record per column. The first three columns of example data describe evidence available by the decision cutoff. The last column is an audit view reconstructed after the report arrived.}
\label{tab:records}
\footnotesize
\setlength{\tabcolsep}{2.5pt}
\begin{tabular}{@{}L{0.21\columnwidth}L{0.152\columnwidth}L{0.152\columnwidth}L{0.2\columnwidth}L{0.203\columnwidth}@{}}
\toprule
 & \multicolumn{2}{l}{Store that sells out} & \multicolumn{2}{l}{Late milestone report} \\
\cmidrule(r){2-3}\cmidrule(l){4-5}
Component & Sales & Demand & Planned & Actual \\
\midrule
Object & \multicolumn{2}{l}{product 1} & \multicolumn{2}{l}{shipment 1} \\
Site or link & \multicolumn{2}{l}{store 1} & \multicolumn{2}{l}{transfer hub} \\
Attribute & sales & demand & \multicolumn{2}{l}{execution status} \\
Value and unit & 40 units & $\geq 40$ units & \multicolumn{2}{l}{arrival} \\
Event time & \multicolumn{2}{l}{day 6} & \multicolumn{2}{l}{hour 28} \\
Time available & day 6 close & day 6 close (bound) & hour 0 (booking) & hour 33 \\
Role & \multicolumn{2}{l}{outcome} & \multicolumn{2}{l}{state} \\
Planned or actual & \multicolumn{2}{l}{actual} & planned & actual \\
Source & \multicolumn{2}{l}{store sales log} & booking plan & carrier milestone report \\
Status & observed & censored & observed & unreported at hour 31 \\
\bottomrule
\end{tabular}
\end{table}

Table~\ref{tab:records} distinguishes an unknown value from an informative bound. When the store sells its last unit at 15:00, the day's sales of 40 establish a lower bound on demand. At hour 31, the freight planner knows the scheduled arrival and sees no report. Only the subsequent report reveals that the shipment arrived on time at hour 28.

Ten sources supply the records. Five are public observational datasets: M5 daily store sales\footnote{\url{https://www.kaggle.com/competitions/m5-forecasting-accuracy}}, BPI Challenge 2019 purchase-order logs\footnote{\url{https://data.4tu.nl/articles/dataset/BPI_Challenge_2019/12715853/1}}, LaDe last-mile deliveries\footnote{\url{https://huggingface.co/datasets/Cainiao-AI/LaDe}}, Olist marketplace orders\footnote{\url{https://www.kaggle.com/datasets/olistbr/brazilian-ecommerce}}, and planned-versus-actual delivery routes\footnote{\url{https://data.mendeley.com/datasets/kkwgfvmtxn/1}}. Five are simulators: Stockpyl\footnote{\url{https://github.com/LarrySnyder/stockpyl}}, SupplyNetPy\footnote{\url{https://github.com/SupplyChainSimulation/SupplyNetPy}}, MABIM\footnote{\url{https://github.com/VictorYXL/ReplenishmentEnv}}, ISOMORPH\footnote{\url{https://github.com/tuhinsahai/ISOMORPH}}, and a discrete-event simulator of shipment milestones that we developed. The sources cover different parts of the operation (Table~\ref{tab:coverage}). The observational datasets mainly cover two modules, supply chain flows and planning and control. Inventory, capacity, and operating state appear in the simulators, which also record quantities that the observational datasets hide, such as true demand and pipeline inventory. Training on all sources exposes the model to relations that each source shows in part.

\begin{table}[t]
\caption{Business modules that the fields of each pretraining source cover. $\bullet$: at least one field maps directly to an attribute of the module. $\circ$: a field supports the module through a derived quantity, such as a delay computed from planned and actual times. Blank: no field maps to the module.}
\label{tab:coverage}
\footnotesize
\setlength{\tabcolsep}{2.4pt}
\begin{tabular}{@{}llcccccc@{}}
\toprule
Source & What it records & \rot{\shortstack[l]{Supply chain\\flows}} & \rot{Inventory} & \rot{\shortstack[l]{Planning and\\control}} & \rot{Capacity} & \rot{Economics} & \rot{\shortstack[l]{Operating\\state}} \\
\midrule
\multicolumn{8}{@{}l}{\textit{Observational datasets}} \\
M5 & daily store sales & $\bullet$ & & & & & \\
BPI Challenge 2019 & purchase-order events & & & $\bullet$ & & $\circ$ & \\
LaDe & parcel deliveries & $\circ$ & & & & & \\
Olist & orders and deliveries & $\bullet$ & & $\bullet$ & & $\bullet$ & \\
Delivery routes & route deviations & $\circ$ & & $\circ$ & & & \\
\addlinespace[2pt]
\multicolumn{8}{@{}l}{\textit{Simulators}} \\
Stockpyl & base-stock chains & $\bullet$ & $\bullet$ & $\bullet$ & & $\bullet$ & \\
SupplyNetPy & networks with outages & $\bullet$ & $\bullet$ & $\bullet$ & $\bullet$ & $\bullet$ & $\bullet$ \\
MABIM & warehouse replenishment & $\bullet$ & $\bullet$ & $\bullet$ & $\bullet$ & $\bullet$ & \\
ISOMORPH & networks with shocks & $\bullet$ & $\bullet$ & $\bullet$ & $\bullet$ & & $\bullet$ \\
Milestone simulator & shipment milestones & $\bullet$ & & $\bullet$ & $\bullet$ & $\bullet$ & $\bullet$ \\
\bottomrule
\end{tabular}
\end{table}

\subsection{Two-stage training}
\label{sec:training}

We train SupplyChain-JEPA in two stages (Figure~\ref{fig:training}). The model reads an episode, a bounded history of one operating unit, as tokens that group available attributes by entity and time or by event. A unit can be a product at a store, a purchase order, a courier day, or a simulated network. The first stage, State pretraining, follows the JEPA principle. A context encoder reads the visible tokens. A target encoder, an exponential moving average of the context encoder, reads the unmasked target window that contains the withheld tokens. The loss uses its outputs at these tokens. A State predictor receives the context representation and a query that names an object, a time, and a horizon. It predicts the target representation. Both encoders are six-layer Transformers of width 384, and the State predictor has three layers. The State encoder is the pretrained context encoder, and the State representation averages its token representations.

Masks withhold contiguous periods, groups of related objects, network neighborhoods, or the remaining milestones of an event sequence. A site-specific outage hides reports from one site for one to six periods. A future query uses only decision-time evidence to predict a target one, two, four, or eight periods ahead. Targets come from recorded values or simulator states available for supervision.

Decisions depend on magnitudes, so the model predicts business quantities, such as demand, inventory, backorders, and lead time, from the predicted representation. It also predicts, for each group of related attributes, a summary of the withheld information. These targets keep operational meaning and magnitude in the representation.

The second stage, Dynamics training, keeps the State encoder frozen and trains a four-layer Dynamics predictor. The predictor receives the State representation of the history up to the decision time, a candidate order, and a horizon. It predicts the representation that the frozen encoder would assign to the future under that order. Training uses paired branches from two inventory simulators, Stockpyl and MABIM. Both branches of a pair start from the same state and use the same pre-sampled demand, which is the method of common random numbers~\cite{glasserman1992common}. The branches differ in the order, for example a small or a large order relative to mean demand. The frozen encoder maps each simulated future to a target, and the predictor learns to predict each future and the difference between the two branches. Additional loss terms penalize a predictor whose output does not change with the order.

Observational records do not enter Dynamics training, because a log does not contain the outcome of an order that was not placed. With the encoder frozen, the current state and the predicted future of each plan lie in one representation space. The trained model has 73.6 million parameters, counting both encoder copies and both predictors. State pretraining uses 37,712 episodes, and Dynamics training uses 2,250 pairs of branches. An application uses the State representation to estimate hidden conditions or to rank risks, and it queries the Dynamics predictor to compare plans. In both cases, a small task model fitted on the application's own labels converts the representation into the business quantity of interest.

\section{Experiments}
\label{sec:experiments}

We evaluate demand recovery, early shipment-risk ranking, and replenishment-plan choice on two datasets and one simulator held out of pretraining. Each application adapts the pretrained model using task labels. Table~\ref{tab:results} compares the resulting system with a conventional baseline given the same decision-time history and labels. For shipment risk and plan choice, a matched random initialization also tests the contribution of the pretrained weights.

The first decision is demand recovery under stockouts. A fresh-grocery retailer sets orders from demand estimates, but hourly sales stop when a product sells out. A model that reads stockout hours as low demand underestimates demand, and orders inherit this bias~\cite{wang2025freshretailnet}. We use FreshRetailNet-50K\footnote{\url{https://huggingface.co/datasets/Dingdong-Inc/FreshRetailNet-50K}}, released by Dingdong, with 50,000 store-product series of hourly sales from 898 stores in 18 cities and hourly stockout annotations. True demand during a stockout is unknown. We therefore hide sales in in-stock periods with masks shaped like real stockouts and score the estimates against the hidden sales. The test stores differ from the stores that supply labels, and the task model uses 3,000 labeled examples, which are 5\% of the labeled pool. The baseline estimates hidden demand from similar past days. Our system represents each period with the frozen State encoder, retrieves similar labeled periods, and adjusts the baseline estimate with a small gradient-boosted model. The weighted absolute percentage error (WAPE) is the sum of absolute errors divided by the sum of hidden sales. Across the three test masks, pooled WAPE falls from 0.4170 for the similar-day baseline to 0.3991 for the full system, a 4.28\% reduction. The system improves under each mask seed.

\begin{table}[t]
\caption{Results on the three applications. Random denotes a matched random initialization with the same downstream procedure. Demand recovery is scored on in-stock sales hidden by stockout-shaped masks.}
\label{tab:results}
\footnotesize
\setlength{\tabcolsep}{3pt}
\begin{tabular*}{\columnwidth}{@{\extracolsep{\fill}}l l r r r@{}}
\toprule
Task & Metric & Baseline & Random & Ours \\
\midrule
Masked in-stock sales & WAPE $\downarrow$ & 0.4170 & --- & 0.3991 \\
Shipment risk & AUPRC $\uparrow$ & & & \\
\quad Confirmation & & 0.3862 & 0.3875 & 0.4122 \\
\quad Adjacent cohort & & 0.4966 & 0.4785 & 0.4774 \\
Plan choice & Regret $\downarrow$ & 9.274 & 6.119 & 5.883 \\
\bottomrule
\end{tabular*}
\end{table}

The second decision is early service-level risk ranking. An airfreight forwarder can review a limited share of shipments for the risk of violating a service-level agreement (SLA). We use the Cargo 2000 log\footnote{\url{https://archive.ics.uci.edu/dataset/382/cargo+2000+freight+tracking+and+tracing}} of 3,942 airfreight processes~\cite{metzger2015predictive}. At the decision time, the latest reports for one critical incoming leg have not arrived, and both systems compute all delay features without them. The baseline is a logistic model that uses a labeled source cohort and 64 labeled shipments from the target cohort. Our system keeps this baseline unchanged. The frozen State encoder represents how a shipment's record changes from an earlier milestone to the decision time. A reference pattern estimated from 47 of these labeled shipments scores this change. The final ranking gives a weight of 80\% to the baseline rank and 20\% to the rank of this score. We also evaluate the same ranking procedure with ten randomly initialized encoders. On the confirmation cohort with joint route and topology shift (126 shipments, 36 violations), AUPRC rises from 0.3862 to 0.4122 ($+6.74\%$), compared with 0.3875 for the mean random encoder. On the adjacent single-shift cohort (1,696 shipments, 468 violations), the baseline ranks better (0.4966 versus 0.4774).

The third decision is the choice between two replenishment plans. A planner has a fixed order quantity and two suppliers, a primary and a secondary one. The regular plan places the full quantity with the primary supplier, and the split plan divides the same total between the two. Each supplier moves between reliable, strained, and disrupted regimes that last several periods, and the planner does not observe these regimes. Capacity reports are the main evidence. In the 16 periods before the decision, they arrive late or not at all, more often when a supplier is impaired. Which plan costs less depends on the hidden regimes. After the decision, the planner places no further orders, and costs accrue over four periods. A historical log shows the outcome of one plan for each state. We therefore evaluate both plans in simulated worlds outside the pretraining data, under common random numbers. Regret is the cost of the chosen plan minus the lower of the two plan costs. The baseline is an ExtraTrees model that predicts the cost of each plan from the observed history. Our system passes the same history through the frozen State encoder, and the Dynamics predictor predicts the future under each plan. A scalar readout converts each future into its cost. We split the simulated worlds five times into disjoint sets for fitting the model, selecting its settings, and reporting results. We adapt the Dynamics predictor and the readout by supervised learning on paired outcomes from the fitting worlds. The adapted system incurs mean regret of 5.883, compared with 9.274 for ExtraTrees, a reduction of 36.6\%. With the same architecture and adaptation procedure, random initialization yields mean regret of 6.119. Pretraining lowers this by 3.85\% on average and yields lower regret on three of five splits.

\raggedbottom
\section{Conclusion and Future Work}
\label{sec:conclusion}

A shared state representation supported demand estimation and shipment-risk ranking, and supplied a starting state for comparing replenishment plans. The shipment result depended on the evaluation cohort. In planning, the matched comparison holds downstream adaptation fixed and estimates the additional gain associated with pretrained weights.

There are several directions for future work. First, we will add attributes for demand forecasts and promotions, and we will represent price as a decision. Second, LLM agents for supply chain operations can use SupplyChain-JEPA. LLM systems already help planners query optimization models and run what-if analyses~\cite{li2023supplychain}. An agent can call SupplyChain-JEPA as the numerical model that estimates hidden state and compares candidate actions. The agent handles language and business rules. Third, paired branches with other actions, such as expediting or transshipment, can extend the Dynamics component to more decisions. The evaluation uses a fixed release of the code\footnote{\url{https://github.com/htizhang/SupplyChain-JEPA}} and pretrained weights\footnote{\url{https://huggingface.co/htizhang/SupplyChain-JEPA}}.

\bibliographystyle{ACM-Reference-Format}
\bibliography{references}

\end{document}